\documentclass{article}

 \usepackage[dblblindworkshop, final]{neurips_2025}
\workshoptitle{Continual and Compatible Foundation Model Updates (CCFM)}

\usepackage[utf8]{inputenc} % allow utf-8 input
\usepackage[T1]{fontenc}    % use 8-bit T1 fonts
\usepackage{hyperref}       % hyperlinks
\usepackage{url}            % simple URL typesetting
\usepackage{subcaption}
\usepackage{booktabs}       % professional-quality tables
\usepackage{graphicx}
\usepackage{amsfonts}       % blackboard math symbols
\usepackage{nicefrac}       % compact symbols for 1/2, etc.
\usepackage{microtype}      % microtypography
\usepackage{xcolor}         % colors
\usepackage{wrapfig}  
\usepackage[most]{tcolorbox}% Required for wrapping text around images
\hypersetup{
    colorlinks=true,
    linkcolor=red,
    filecolor=magenta,
    urlcolor=blue,
    citecolor=purple,
    pdftitle={Overleaf Example},
    pdfpagemode=FullScreen,
    }

\newcommand{\fs}{$f_{S}$}
\newcommand{\ft}{$f_{T}$}
\newcommand{\fm}{$f_{M}$}

\newcounter{qcounter}
\title{When Data Falls Short: Grokking Below the Critical Threshold}

\author{%
  Vaibhav Singh$^{1,2}$ \hspace{2pt}   \quad Eugene Belilovsky$^{1,2}$ \quad  Rahaf Aljundi$^{3}$\\
  $^1$Concordia University \quad $^2$Mila \quad $^3$Toyota Motor Europe \\
}

\begin{document}

\maketitle

\begin{abstract}
In this paper, we investigate the phenomenon of \textit{grokking}, where models exhibit delayed generalization following overfitting on training data. We focus on data-scarce regimes where the number of training samples falls below the critical threshold, making grokking unobservable, and on practical scenarios involving distribution shift. We first show that Knowledge Distillation (KD) from a model that has already grokked on a distribution ($p_1$) can induce and accelerate grokking on a different distribution ($p_2$), even when the available data lies below the critical threshold. This highlights the value of KD for deployed models that must adapt to new distributions under limited data. We then study training on the joint distribution ($p_1$, $p_2$) and demonstrate that while standard supervised training fails when either distribution has insufficient data, distilling from models grokked on the individual distributions enables generalization. Finally, we examine a continual pretraining setup, where a grokked model transitions from $p_1$ to $p_2$, and find that KD both accelerates generalization and mitigates catastrophic forgetting, achieving strong performance even with only 10\% of the data. Together, our results provide new insights into the mechanics of grokking under knowledge transfer and underscore the central role of KD in enabling generalization in low-data and evolving distribution settings.
    
\end{abstract}
\section{Introduction}
\label{sec:intro}

Generalizing across varying data distributions remains a core challenge in machine learning, as standard training often fails under distribution shifts or data scarcity~\citep{van2019three, fang2020rethinking, singh2024wake,singh2024controlling, liang2024comprehensive}. The phenomenon of \textit{grokking} \citep{power2022grokking} sheds light on this problem, showing how models can suddenly generalize after prolonged overfitting \citep{Arpit2017ACL}. Explanations range from implicit regularization, such as weight decay~\citep{barak2022hidden, nanda2023progress}, to training dynamics that enable generalization even at zero loss~\citep{kumar2024grokking, lyu2024dichotomy, Ishida2020DoWN}. A common finding is that grokking occurs only when training data exceeds a \textit{critical threshold}~\citep{power2022grokking, critical_data}.

Building on these findings, we explore grokking in scarce data regimes, under distribution shift. Specifically, we ask:\textbf{ Can a grokked model be leveraged to \textit{\textbf{train}} another model on a different distribution?} To test this, we train a one-layer Transformer~\citep{Vaswani2017AttentionIA} on $p_1$ as a Teacher (\ft) and distill its knowledge into a Student (\fs) on $p_2$. We find that \fs\ not only groks on $p_2$ but also requires fewer steps under distillation. This enables faster adaptation when $p_2$ has limited data, demonstrating the practical value of pre-grokked models for distribution shift, continual learning, multi-task learning, and domain generalization.

We also show that generalization does not depend on smaller weight norms or weight decay. While prior work linked grokking to a cleanup phase driven by weight decay~\cite{nanda2023progress, liu2022omnigrok, varma2023explaining}, our experiments refute this claim. Consistent with recent findings~\cite{prieto2025grokking}, we find that weight decay mainly mitigates floating-point errors. Grokking occurs even with zero weight decay and increasing weight norms, ruling out these factors as primary explanations.

We further ask: \textbf{Is grokking possible when training data falls \textit{below} the \textbf{\textit{critical threshold}} ?} Our results show that distillation from the Teacher model (\ft) not only reduces the steps required for grokking but also enables it in regimes where data is below the critical size. The critical data size, as defined in~\citep{liu2022omnigrok, varma2023explaining, critical_data}, is the minimum amount of data below which generalization does not occur. Our study further demonstrates that KD helps mitigate forgetting when models adapt to new domains. Together, these contributions outline a practical framework for building efficient and adaptable learning systems.

\begin{figure}[t]
    \centering
    \begin{subfigure}{0.4\textwidth}
        \centering
        \includegraphics[width=\linewidth]{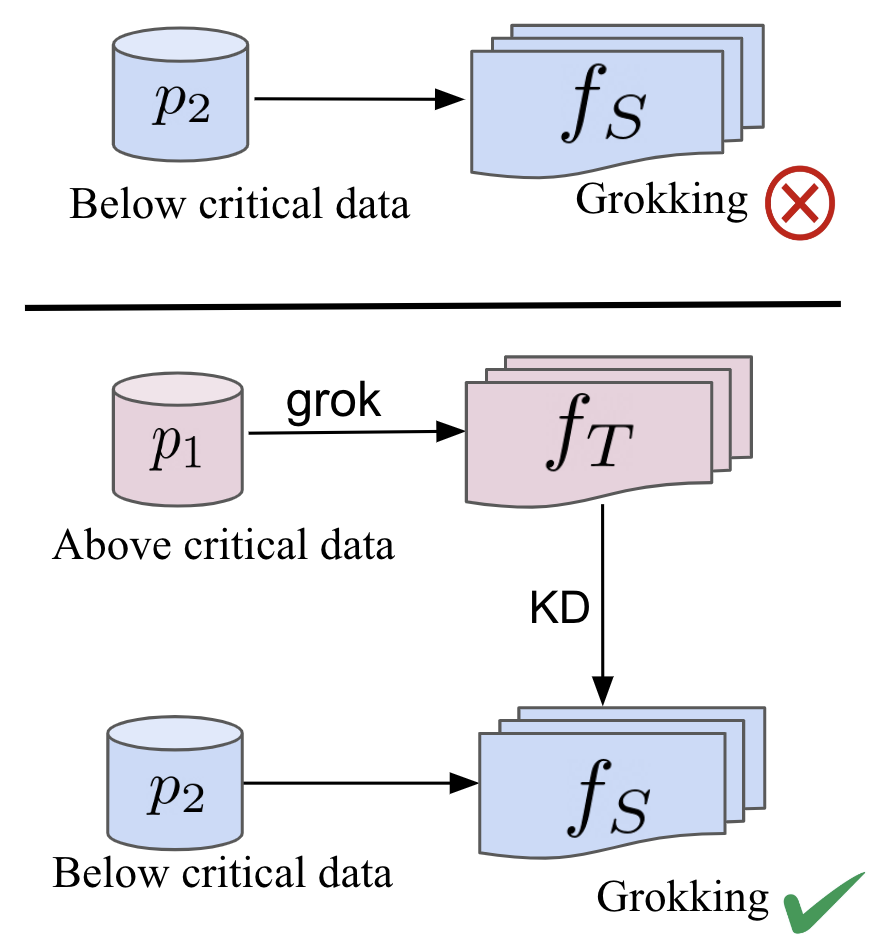}
        \caption{Grokking a model on $p_2$ using KD, which otherwise fails to generalize.}
        \label{fig:scheme_11}
    \end{subfigure}
    \hfill
    \begin{subfigure}{0.4\textwidth}
        \centering
        \includegraphics[width=\linewidth]{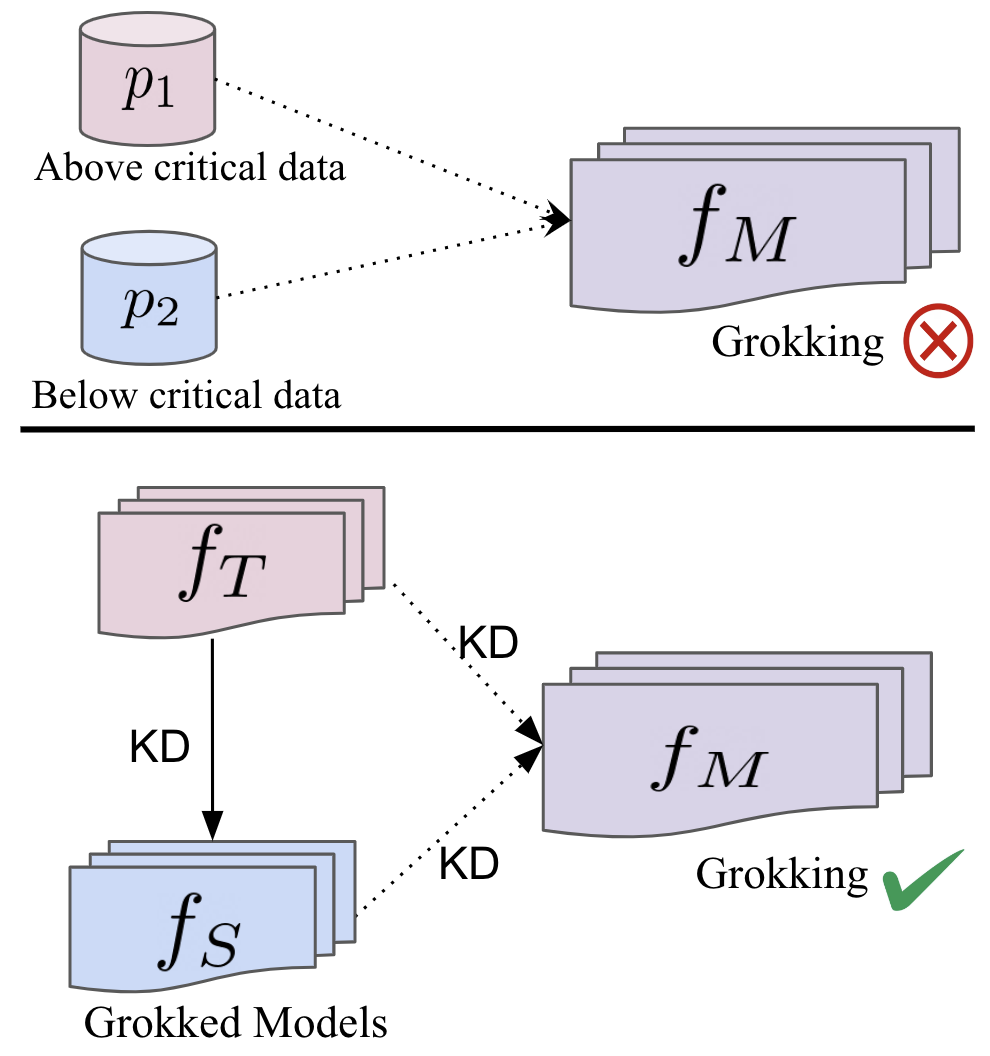}
        \caption{Distilling from multiple grokked models \ft, \fs\ yields grokking on a larger model \fm\ below critical data.}
        \label{fig:scheme_22}
    \end{subfigure}
    
    \caption{\small \autoref{fig:scheme_11} shows that \fs groks below the critical data size when trained via KD from a grokked model \ft, whereas training from scratch fails. In \autoref{fig:scheme_22}, a larger model \fm trained jointly on $p_1$ and $p_2$ cannot generalize when either dataset is below the threshold. Distilling from the smaller grokked models \fs and \ft, however, enables \fm\ to grok and generalize effectively even under scarce data.}
    \label{fig:scheme_diag}
 \end{figure}

\section{Related Work}
\label{sec:related}

\textbf{Grokking} was first identified in algorithmic tasks by \citep{power2022grokking}. Subsequent work has aimed to explain this phenomenon. \citep{nanda2023progress} reverse-engineered a grokked modular-addition transformer and found it learns a composition of trigonometric and inverse-trigonometric functions. \citep{merrill2023tale} attributed grokking to competition between sparse (generalizing) and dense (memorizing) subnetworks. From a theoretical lens, \citep{rubin2024grokking} framed grokking as a first-order phase transition in feature learning, while \citep{levi2024grokking} provided analytical solutions for loss and accuracy dynamics in linear networks. Studying polynomial regression, \citep{kumar2024grokking} linked grokking to a shift from lazy to rich learning. \citep{lyu2024dichotomy} further suggested that the sharp test-accuracy jump arises from differing implicit biases in early vs.\ late training.

Grokking has also been observed in practical settings, e.g., CNNs trained on CIFAR-10~\citep{humayun2024deep, cifar10}. \citep{humayun2024deep} describe delayed robustness, where models eventually grok adversarial examples long after interpolation. Early prediction of grokking has been attempted using Fourier spectral signatures~\citep{notsawo2023predicting}. It has been linked to slow formation of useful representations within a “Goldilocks zone” between memorization and confusion~\citep{liu2022towards}, and to gradual amplification of structured weights followed by removal of memorized components~\citep{nanda2023progress}. Other explanations include hidden SGD-driven amplification of a Fourier gap~\citep{barak2022hidden} and the “Slingshot mechanism,” where training cycles between stable and unstable phases~\citep{thilak2022slingshot}.

\textbf{Relationship to Dataset Size:} Circuit-efficiency analysis shows that generalization is slower but more efficient, implying a critical data size below which models memorize rather than generalize~\citep{varma2023explaining}. Training below this threshold yields semi-grokking, and fine-tuning grokked models on such small data can cause “ungrokking.” Regularization has been proposed to correct training-sample errors~\citep{doshi2023grok}, and loss-landscape analysis links grokking to data size, weight decay, and representation learning~\citep{liu2022omnigrok}.

\textbf{Accelerating Grokking:} Several methods speed up grokking: gradient decomposition and amplification~\citep{lee2024grokfast}, lottery-ticket approaches~\citep{minegishi2023bridging}, transferring embeddings from a weaker to a stronger model~\citep{xu2025let}, and replacing softmax with stable-max~\citep{prieto2025grokking}. In contrast, our method removes the phase transition without extra data or redundant pretraining, and to our knowledge is the first to accelerate grokking in data-scarce settings under distribution shift.

% \textbf{Knowledge Distillation(KD): } Knowledge distillation ~\cite{hinton2015distilling} is a widely used technique for model compression~\cite{sun2019patient, sarfraz2021knowledge, mishra2017apprentice}, building more efficient neural network families ~\citep{huang2017densely, singh2024controlling, singh2024wake}, quantizing existing networks to
% use fewer bits for weights and activations ~\cite{wu2016quantized} and distilling knowledge from larger networks into smaller ones~\citep{tung2019similarity}. The method involves training a smaller student model to replicate the behavior of a larger teacher model. This approach has been applied successfully in various domains, including natural language processing and computer vision. Our work builds on this foundation by focusing on task-level knowledge transfer in algorithmic tasks with varying data distributions. 

\section{Experimental Setup}
\label{appendix:loss_formulation}

We trained a decoder only transformer to perform experiments on algorithmic tasks of the form  ($(a @ b) \% P$), where $@$ represents operator for any of the binary operations. In this work, we focus on addition and subtraction tasks following previous studies~\citep{nanda2023progress, varma2023explaining, liu2022omnigrok, power2022grokking, liu2022towards} which consistently report grokking on these tasks. The model input is $[a, b, @, P]$, and the output $c$ is read from the final token $P$. In our experiments, each arithmetic modulo-$P$ task is denoted as $p_1$ for a specific prime $P$. A distribution shift is introduced by changing the modulus while keeping the operator fixed. For example, in addition modulo $P$, the task $(a+b)\%P$ with $P=P_1$ is referred to as distribution $p_1$, while $(a+b)\%P$ with $P=P_2 \neq P_1$ is denoted $p_2$. Our results remain consistent regardless of the choice of $P_1$ and $P_2$.
   
   We begin training with 35\% of the dataset to first observe grokking, as demonstrated in prior work~\citep{power2022grokking, liu2022omnigrok}. We then progressively reduce the data fraction to 30\%, 25\%, and 10\%. For algorithmic addition and subtraction tasks, prior studies define the critical data size to be around 25\% of the dataset~\citep{varma2023explaining, critical_data}. Consistent with this, our observations show that grokking does not occur when 25\% or less of the data is available, confirming that 25\% marks the critical threshold below which generalization becomes impossible.
We utilize StableMax Cross Entropy \citep{prieto2025grokking} since cross entropy with softmax function causes numerical instability, given as: 
   \begin{equation}
       L_{\mathrm{StCE}}(\theta)=\mathbb{E}_{(x, y) \sim \mathcal{D}}\left[-\log \text{StableMax} (f_S^y(x ; \theta))\right]
   \end{equation}
   where $f_S(\cdot ; \theta)$, is the  Student Network: parameterized by $\theta$ and StableMax($x$) = Softmax($g(x_i)$) where $g(x_i)$ is defined as 

   \begin{equation}
    g(x)= \begin{cases}\log (x+1) & \text { if } x \geq 0 \\ -\log (-x+1) & \text { if } x<0\end{cases}
    \end{equation}
   We observe that the usage of StableMax \citep{prieto2025grokking} already gives a prior speedup in inducing grokking, needing around 6000 iterations to grok, which otherwise would have been in order of $1e4$ \citep{liu2022omnigrok, power2022grokking, nanda2023progress} 

   For knowledge distillation, we use Kullback-Leibler (KL) Divergence Loss: 
   \begin{equation}
       L_{\mathrm{KL}}(\theta)=\mathbb{E}_{x \sim \mathcal{D}_X}\left[D_{\mathrm{KL}}\left(q_T(x) \| q_S(x ; \theta)\right)\right]
   \end{equation}
   where $D_{\mathrm{KL}}(p \| q)=\sum_{i=1}^K p_i \log \left(\frac{p_i}{q_i}\right)$. This takes softened outputs as $q_T(x)=\operatorname{softmax}\left(\frac{f_T(x)}{t}\right)$, and $q_S(x ; \theta)=\operatorname{softmax}\left(\frac{f_S(x ; \theta)}{t}\right)$ where $f_T:$, represents the Teacher model, and $t>0$ is the Temperature used to soften probabilities. 
   
   The total distillation loss is therefore realised as: 
   \begin{equation}
       L(\theta)=(1-\alpha) L_{\mathrm{CE}}(\theta)+\alpha L_{\mathrm{KL}}(\theta)
   \end{equation} 
   where $\alpha$ controls the proportion of each loss component.

   For demonstrating the efficacy of our distillation method and to negate the dependency of weight norm and weight decay theories, we compare both Adam without weight decay \& AdamW(with weight decay) optimizer ~\citep{loshchilov2017decoupled} with a learning rate $\gamma=1e-3$. For AdamW we set the weight decay parameter $\lambda=1$. We perform 15,000-30,000 epochs of training with a batch size of 2048 on NVIDIA V100 GPU.

\section{Accelerating Grokking through Knowledge Distillation (KD)}
   \label{sec:grok_with_kd}
    \begin{figure}[t!]
      \centering
      \begin{subfigure}{0.49\textwidth}
          \centering
          \includegraphics[width=\linewidth]{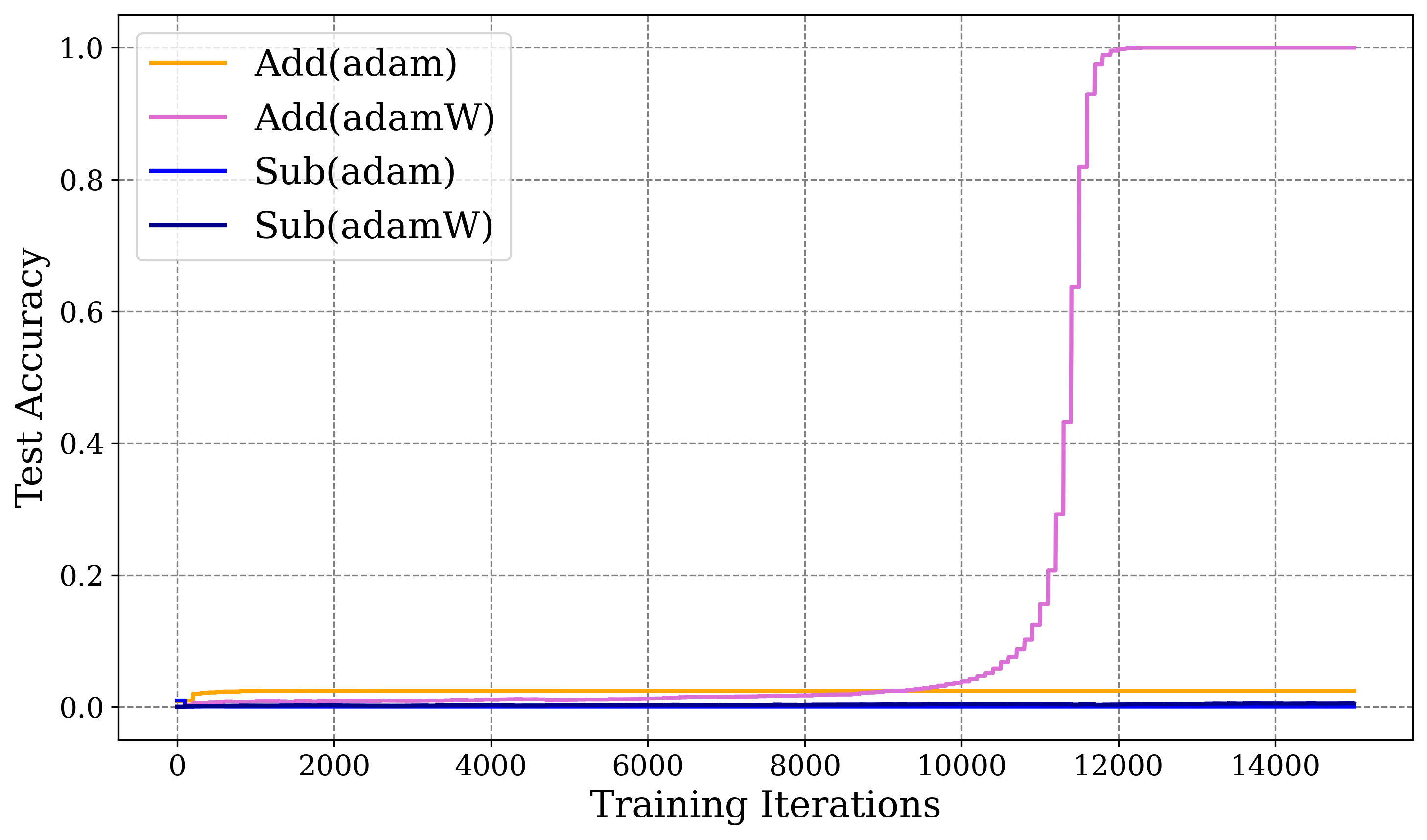}
          \caption{Training on $p_2$ without KD (f = 30\%)}
          \label{fig:vg113}
      \end{subfigure}
      \hfill
      \begin{subfigure}{0.49\textwidth}
          \centering
          \includegraphics[width=\linewidth]{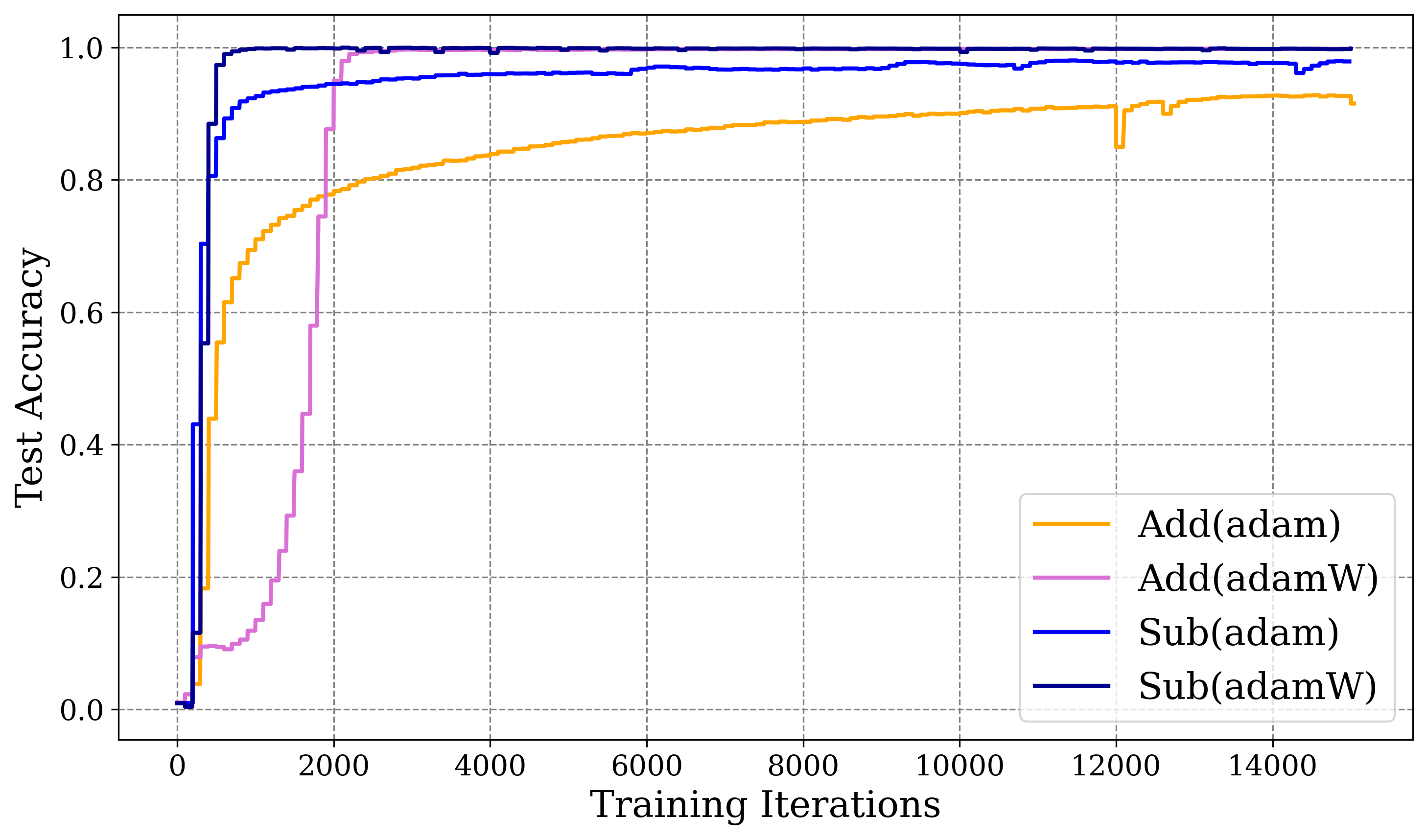}
          \caption{Training on $p_2$ with KD from $p_1$ (f = 30\%)}
          \label{fig:vg107}
      \end{subfigure}
      \caption{\autoref{fig:vg107} shows the effectiveness of KD \textbf{irrespective of the optimizer choice} for both addition and subtraction modulo task. In \autoref{fig:vg113} typical grokking phenomena on distribution $p_2$ on $30\%$ of training data (denoted as f), without KD. We observe that weight decay is helpful in showing grokking but its not the only underlying cause. When trained with Adam, grokking is not observed for both tasks when trained for 15000 iterations. This concurs with \citep{power2022grokking}. However \autoref{fig:vg107} demonstrates a $Student$ model trained on a different distribution $p_2$ with same fraction, but now with KD from the $Teacher$ model trained on $p_1$ displays accelerated generalization irrespective of the optimizer choice.}
      \label{fig:grokking_optim}
  \end{figure}
KD has been shown to provide multiple benefits in improving training dynamics. \cite{pmlr-v139-menon21a} provided a statistical perspective on distillation, that providing the true class-probabilities from the teacher model can lower the variance of the student objective, and thus improve performance. Further \cite{pmlr-v97-phuong19a} provides a generalization bound that establishes fast convergence of the expected risk of a distillation-trained linear classifier. In \citep{boixadsera2024theorymodeldistillation, ben2011learning}, a theoretical framework is given to analyze model distillation into decision trees through PAC-learning statistical theory. They show that if teacher model \ft is perfectly distillable into a student model $f_S$, then with a probability of at least $1-\delta$, \fs generalizes with training samples no greater than $\left\lceil \frac{\log(1/\delta)}{\epsilon} \right\rceil$. It can be inferred from these studies~\citep{tang2020understanding, Cho_2019_ICCV, Yuan_2020_CVPR}  that KD brings the following advantages towards training dynamics, 
   \begin{itemize}
       \item \textbf{Regularization Effect through Label Smoothing:} KD smooths the labels, which acts as a regularizer and prevents overfitting.
       \item \textbf{Domain Knowledge Injection:} The teacher model imparts class relationships that shape the geometry of the student’s logit layer. 
       \item \textbf{Instance-Specific Knowledge:} The teacher adjusts the student model’s per-instance gradients based on the difficulty of each sample, facilitating more effective learning.
   \end{itemize}

  %%%%%%%%%%%%%%%%%%%%%%%%%%%%%%%%%%%%%%%%%%%%%%

%%%%%%%%%%%%%%%%%%%%%%%%%%%%%%%%%%%%%%%%%%%%%%%
\begin{figure}[t]
    \centering
    \begin{subfigure}{0.46\textwidth}
        \centering
        \includegraphics[width=\linewidth]{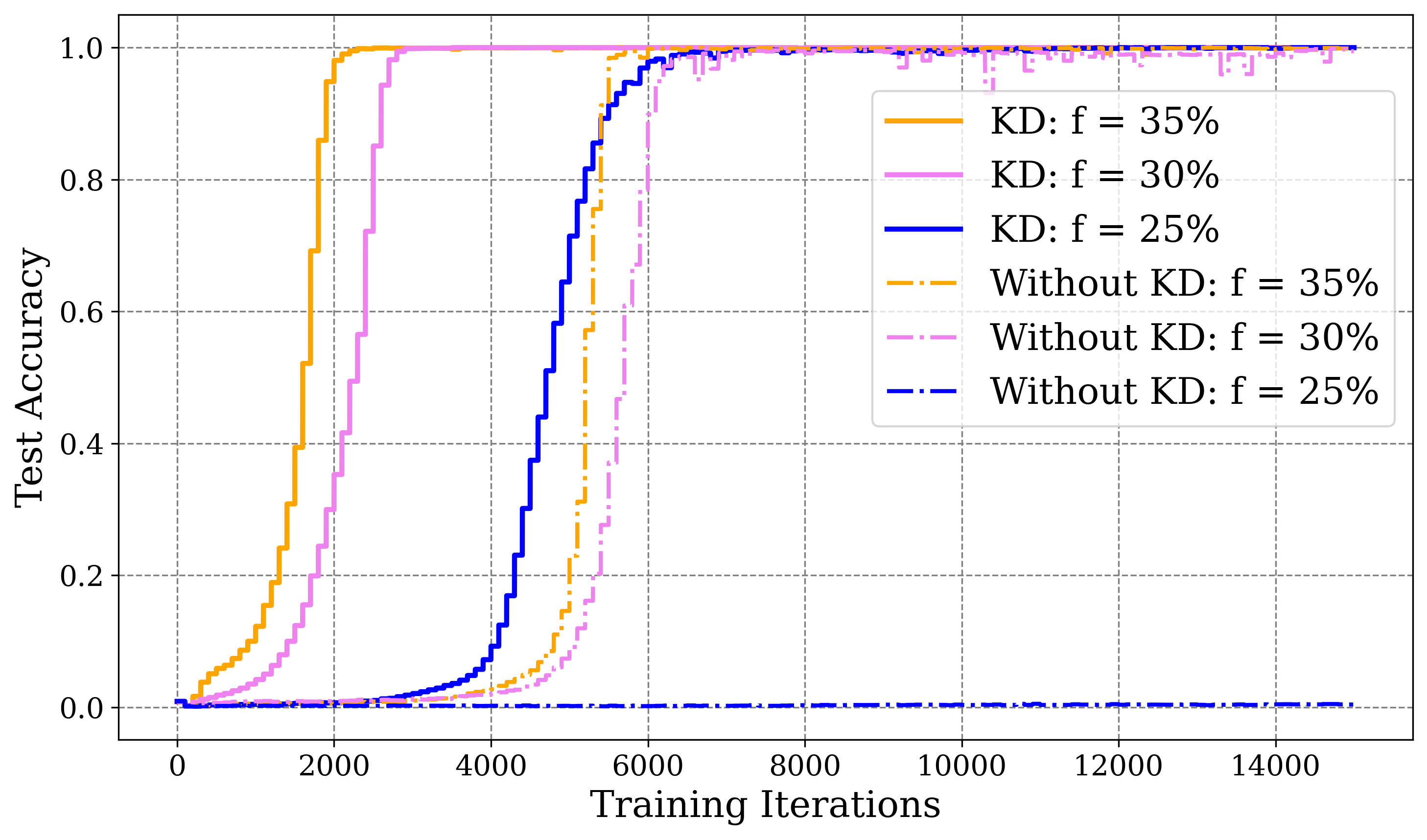}
        \caption{Accelerated Grokking on $p_2$ for Task: $(a+b)\%107$}
        \label{fig:Addition_107_kd}
    \end{subfigure}
    \hfill
    \begin{subfigure}{0.46\textwidth}
        \centering
        \includegraphics[width=\linewidth]{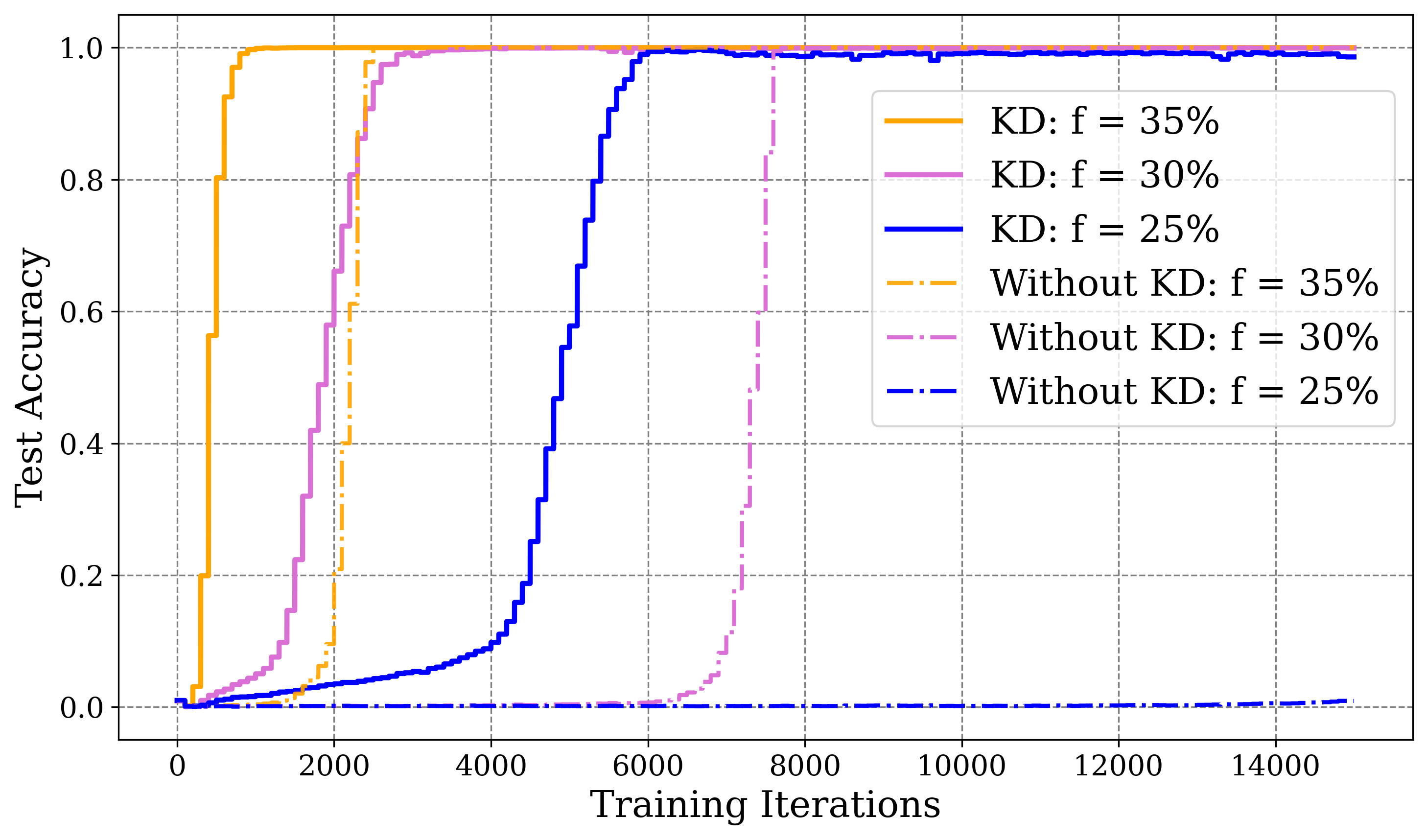}
        \caption{Accelerated Grokking on $p_2$ for Task: $(a-b)\%107$}
        \label{fig:subtraction_107_kd}
    \end{subfigure}
    \caption{\small Dashed lines in\autoref{fig:Addition_107_kd} and \autoref{fig:subtraction_107_kd} show typical grokking on $p_{2}$ $(P=107)$ with different training fractions ($f$). Training from scratch below $30\%$ shows no grokking. With KD from a grokked model on $p_{1}$ $(P=113)$, grokking is accelerated and occurs with as little as $25\%$ of $p_{2}$. Distillation is applied to probability outputs from the operator token, enabling generic operator-level representations rather than $P$-specific ones. }
    \label{fig:acc_grokking}
\end{figure}
%%%%%%%%%%%%%%%%%%%%%%%%%%%%%%%%%%%%%%%%%%%%%%%

We first grok a 1 layer Transformer model with $35\%$ of training data for modular addition and subtraction tasks ($(a \pm b) \% P$) with $P=113$ (Choice of $P$ was aritrary). We call this as data distribution $p_1$. This model will act as the teacher model (\ft). Next we train another model on a distribution $p_2$, by modifying the modulo prime ($P=107$), and compare the impact of KD under different fractions for $p_2$ on both these tasks. As seen in \autoref{fig:acc_grokking}, KD significantly accelerates the grokking process for both tasks, even in scenarios where the proportion of training data is below critical dataset size. This observation is independent of the optimizer used, as shown in \autoref{fig:grokking_optim}. This demonstrates a practical utility of grokked models, illustrating their effectiveness in training models on varying distributions through KD. It is important to note that distillation occurs on the probability outputs from the operator token rather than the $P$ token. This approach aims to learn generic operator-level representations instead of task-specific representations, which would depend on the choice of $P$.

\begin{figure}[t!]
    \centering
    \begin{subfigure}{0.46\textwidth}
        \centering
        \includegraphics[width=\linewidth]{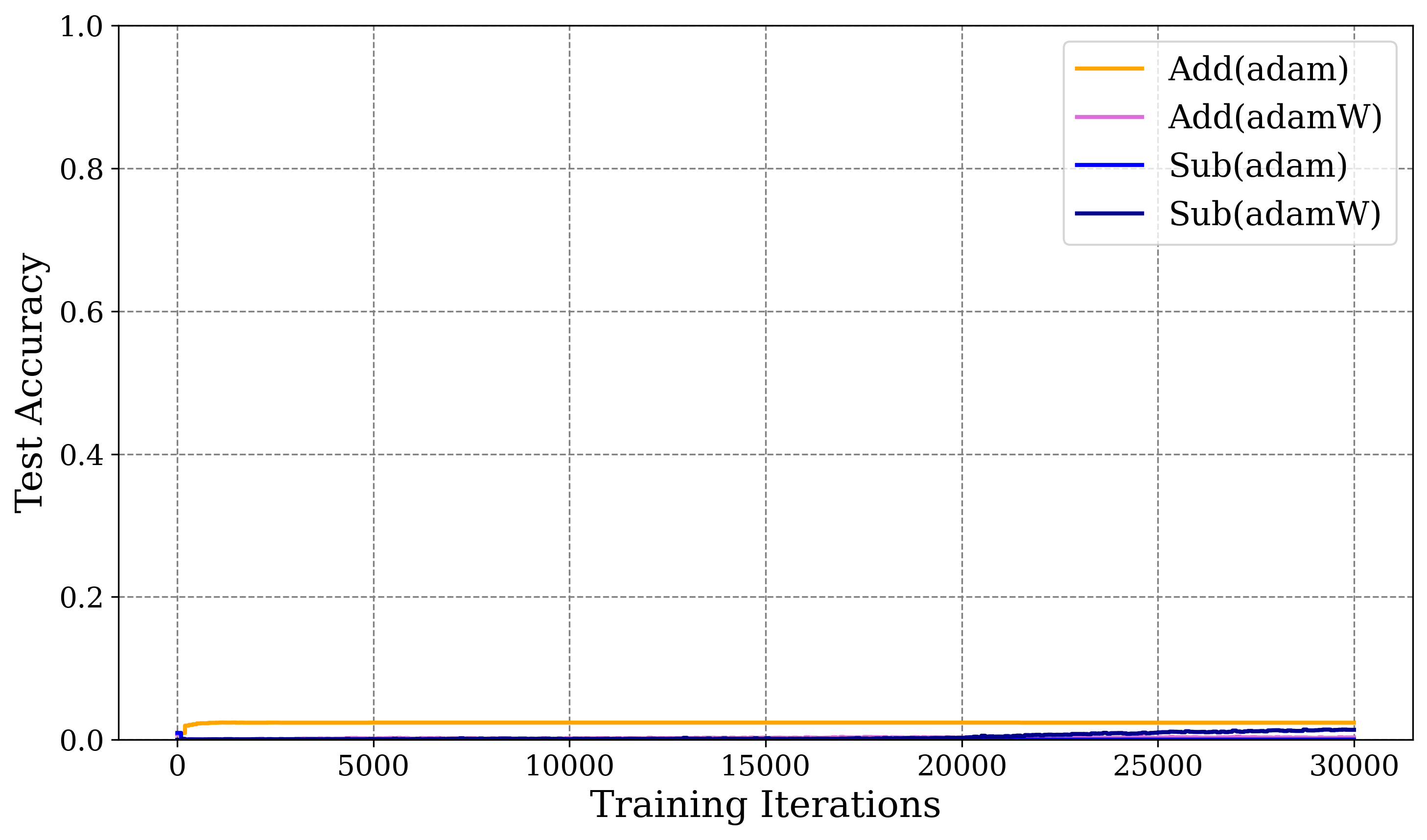}
        \caption{Without KD on $20\%$ of $p_2$}
        \label{fig:without_kd_single}
    \end{subfigure}
    \hfill
    \begin{subfigure}{0.46\textwidth}
        \centering
        \includegraphics[width=\linewidth]{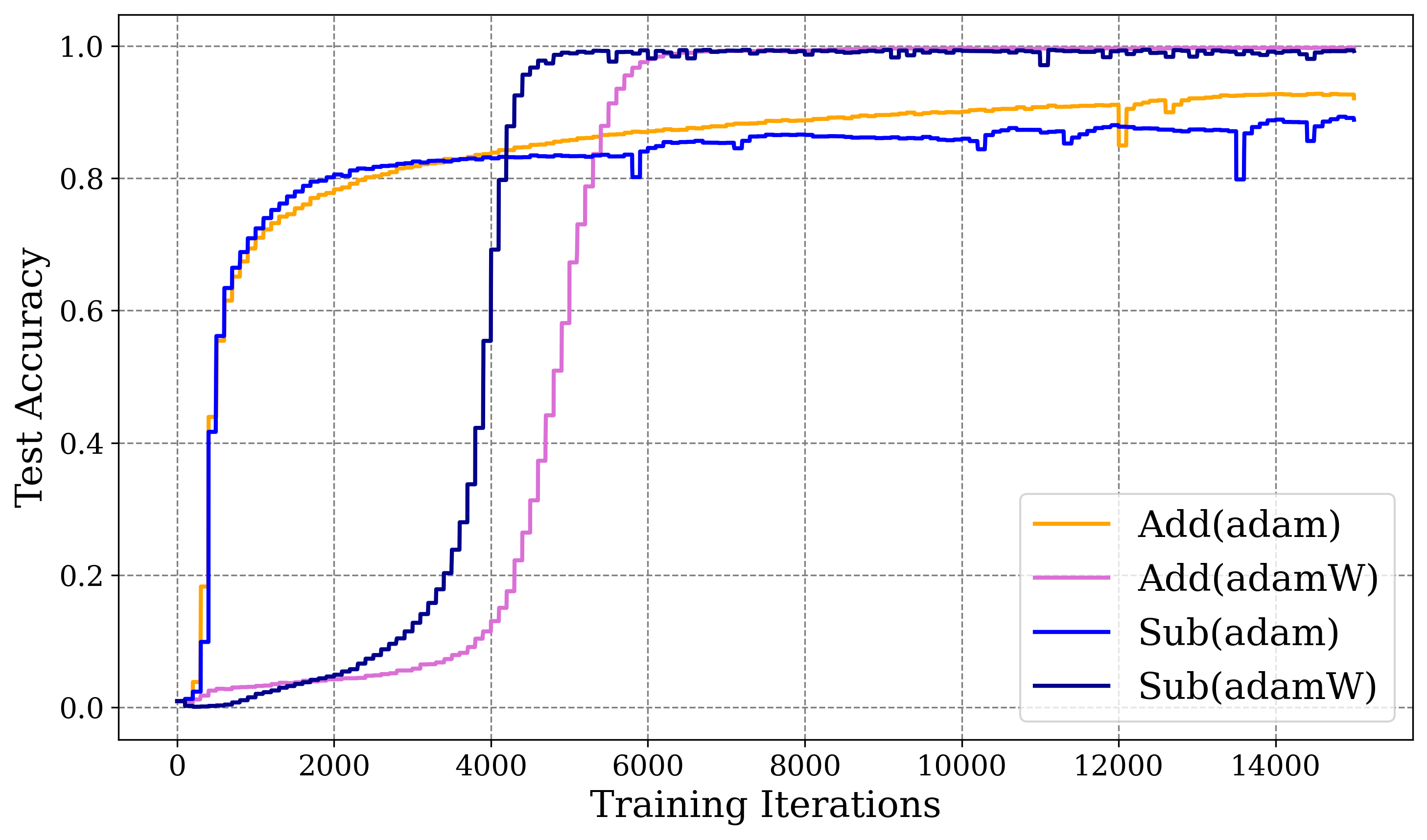}
        \caption{With KD on $20\%$ of $p_2$}
        \label{fig:with_kd_single}
    \end{subfigure}
    \caption{\small \autoref{fig:without_kd_single} demonstrates that its impossible to observe grokking when the data fraction goes below a certain critical threshold($20\%$.), even with 2X iterations (30,000) In such a case, the model does not learn anything regardless of the optimizer. In \autoref{fig:with_kd_single}, it can be clearly seen that with KD, grokking is observed for all tasks, even without weight decay. However we notice that weight decay helps in achieving a better generalisation.}
    \label{fig:kd_frac_0.2}
\end{figure}
  
Building on these observations, we ask: \textit{\textbf{Can KD enable generalization below the critical data threshold?}} To test this, we repeat the experiments with only $20\%$ of the data. Without KD, grokking does not occur even after 30,000 iterations, regardless of weight decay. In contrast, with KD, generalization emerges rapidly at this reduced data fraction (\autoref{fig:kd_frac_0.2}). Notably, the weight norm continues to increase (more details on weight norm given in \ref{appendix:l2norm}), reinforcing our earlier claim that neither weight decay nor decreasing weight norms are essential for grokking. These results underscore the value of a grokked Teacher model (\ft) in data-scarce settings. KD not only accelerates grokking but also makes it possible below the critical threshold, highlighting its practical utility for efficient training under limited data and shifting distributions. 
   
%    These findings highlight the critical role of a grokked Teacher model (\ft), especially in data-scarce environments where the available training data falls below the threshold necessary for grokking or any generalisation. By leveraging a grokked Teacher model through KD, we not only accelerate the grokking process but also extend its applicability to situations with limited data. This demonstrates the practical utility of grokked models in facilitating efficient training across varying data distributions, thereby offering a robust solution for scenarios where data is constrained. 
   
%    We further find that grokking occurs even as the parameter weight norm (L2) increases during training with Adam, challenging the view that decreasing weight norms are essential~\citep{liu2022omnigrok, nanda2023progress, varma2023explaining}. Detailed results and analysis are provided in Appendix~\ref{appendix:l2norm}.

 \section{Leveraging Grokked Models for Distillation and Continual Transfer}
\label{sec:dist-mult-gm}

\begin{figure}[t!]
       \centering
       \begin{subfigure}{0.46\textwidth}
           \centering
           \includegraphics[width=\linewidth]{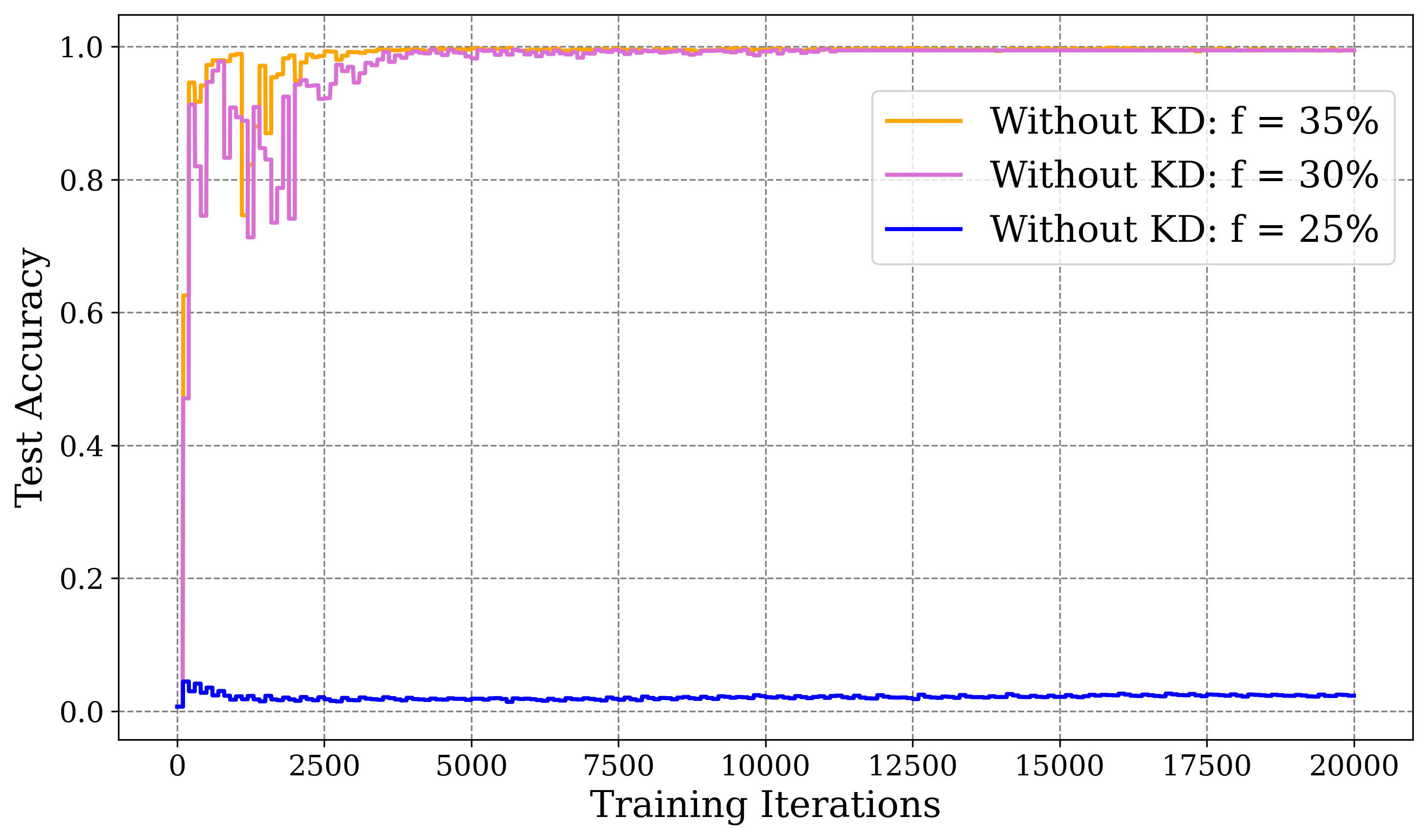}
           \caption{without KD}
           \label{fig:without_kd}
       \end{subfigure}
       \hfill
       \begin{subfigure}{0.46\textwidth}
           \centering
           \includegraphics[width=\linewidth]{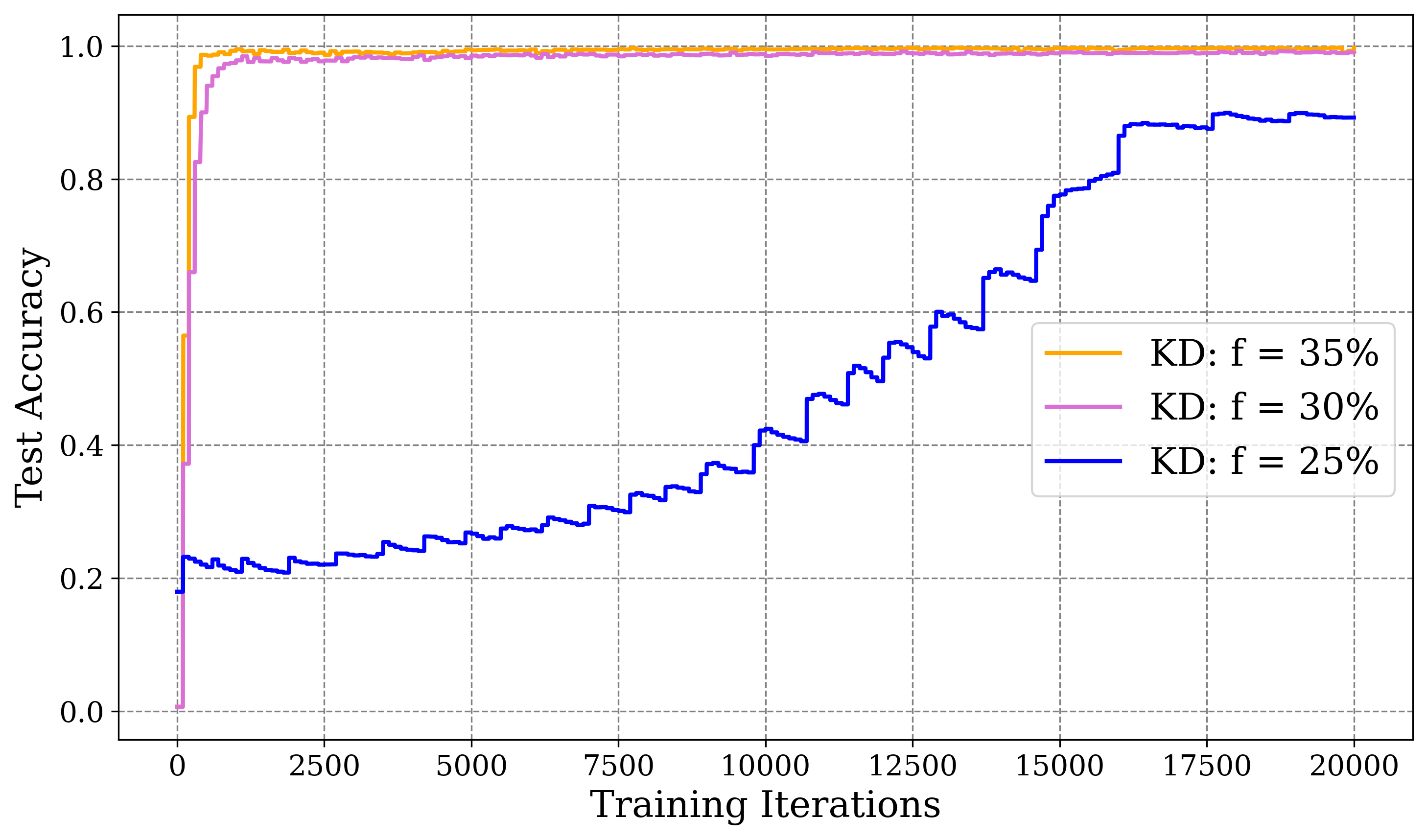}
           \caption{With only KD, without entropy minimization}
           \label{fig:with_kd}
       \end{subfigure}
       
       \caption{\small Performance comparison of training strategies for a larger transformer model \fm\  on distributions of $p_1$(35\%) and different fractions $(0.35, 0.3, 0.25)$ of $p_2$. In the Joint Training regime (\autoref{fig:without_kd}), the model fails to generalize via cross-entropy when data from either distribution falls below the critical threshold. In contrast, training solely with distillation enables grokking even with $25\%$ of $p_2$ (\autoref{fig:with_kd}). At this low fraction, generalization does not reach unity due to the imperfect $f_{p_2}$ trained under data scarcity, while for $0.35$ and $0.3$ fractions, generalization is rapid with no grokking.}
       \label{fig:multiple}
   \end{figure}

We extended our study by checkpointing grokked models on fractions $(0.35, 0.3, 0.25)$ of $p_2$ trained via distillation (Section~\ref{sec:grok_with_kd}). The model grokked on $p_1$ with $35\%$ data is denoted $f_{p_1}$, and the KD-trained models on $p_2$ as $f_{p_2}$. Our goal was to train a new transformer (\fm) that generalizes across both $p_1$ and $p_2$. In joint training, \fm\ failed to generalize when $p_2$ was below the critical size, showing that scarcity in any distribution limits learning. In contrast, training \fm\ solely via KD from $f_{p_1}$ and $f_{p_2}$ enabled simultaneous generalization, even when either distribution had limited data (\autoref{fig:multiple})
    
%    Remarkably, \fm\ exhibited grokking behavior \textbf{\textit{only when trained via KD}}, even when either $p_1$ or $p_2$ was below the critical data size, successfully generalizing despite the limited data. This demonstrates that KD over the joint distribution $(p_1, p_2)$ provides a more informative signal than training with ground-truth labels. KD-enabled training allows grokking to emerge even when the data is below the critical size. Notably, this effect holds true even when the grokked teacher model $f_{p_2}$ was trained on a similarly small fraction of $p_2$ data, but with distillation from $f_{p_1}$.

Remarkably, \fm\ exhibited grokking \textbf{\textit{only when trained via KD}}, generalizing even when $p_1$ or $p_2$ was below the critical size. This shows that KD over the joint distribution $(p_1, p_2)$ provides a stronger learning signal than ground-truth labels, enabling grokking under limited data. Notably, the effect persists even when the teacher $f_{p_2}$ itself was trained on a small fraction of $p_2$, distilled from $f_{p_1}$.
     
% Building on recent advancements in continual pretraining methodologies~\citep{ke2022dgs}, we conducted a comprehensive experiment to assess the efficacy of transitioning a previously grokked model from $p_1$ to $p_2$. Specifically, we examined the role of knowledge distillation (KD) in mitigating catastrophic forgetting during this transition. Our experimental setup involved initializing pretraining with a model that had already achieved generalized performance on $p_1$ through grokking. We then continued pretraining on $p_2$ under two distinct conditions: with and without the application of KD.

Building on recent work in continual pretraining~\citep{ke2022dgs}, we evaluated the transition of a grokked model from $p_1$ to $p_2$, focusing on the role of knowledge distillation (KD) in mitigating catastrophic forgetting. A model grokked on $p_1$ was further pretrained on $p_2$ under two conditions: with and without KD.
   
\begin{figure}[h!]
       \centering
       \begin{subfigure}{0.46\textwidth}
           \centering
           \includegraphics[width=\linewidth]{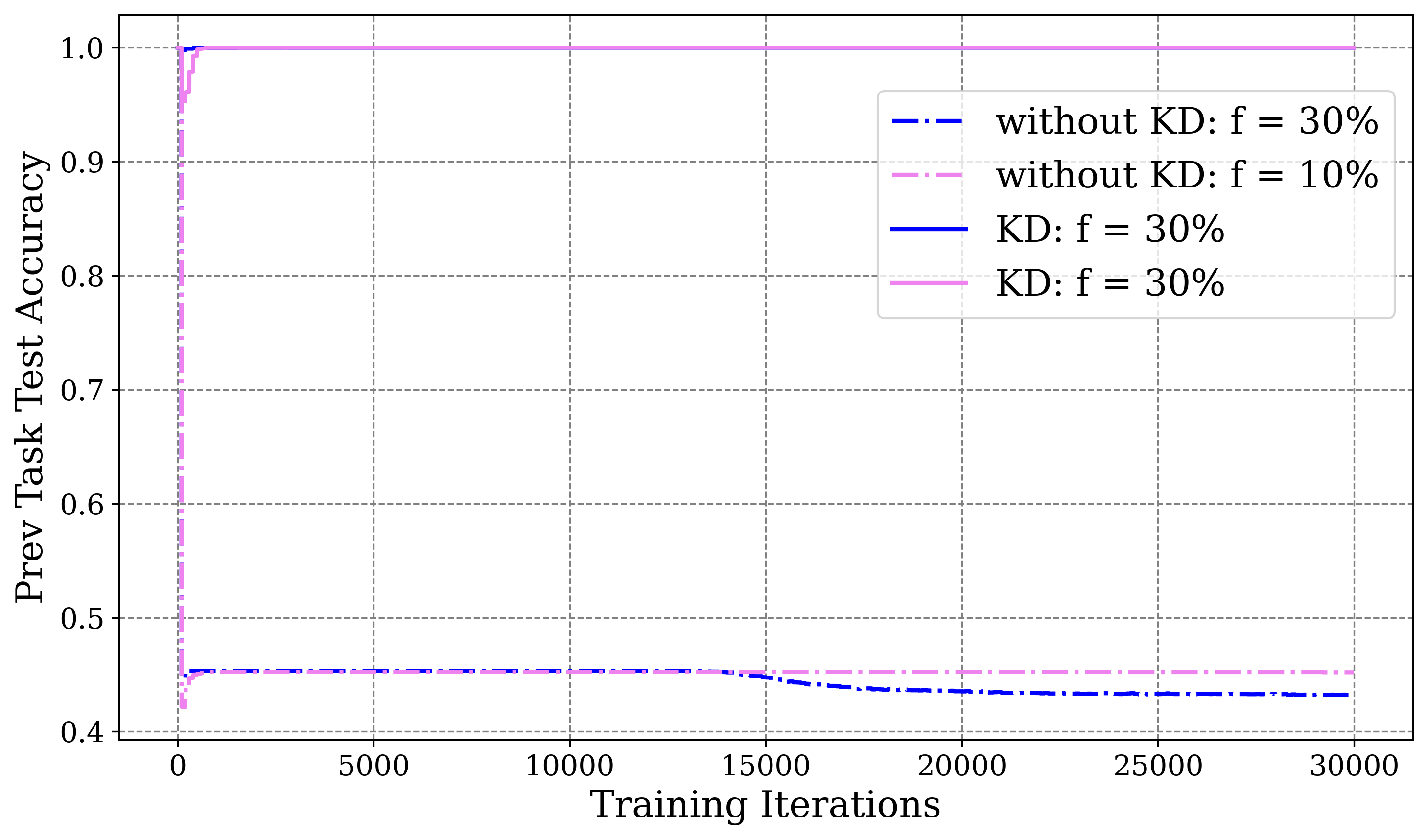}
           \caption{Previous Task Accuracy for different fractions of data, with and without KD.}
           \label{fig:pta}
       \end{subfigure}
       \hfill
       \begin{subfigure}{0.46\textwidth}
           \centering
           \includegraphics[width=\linewidth]{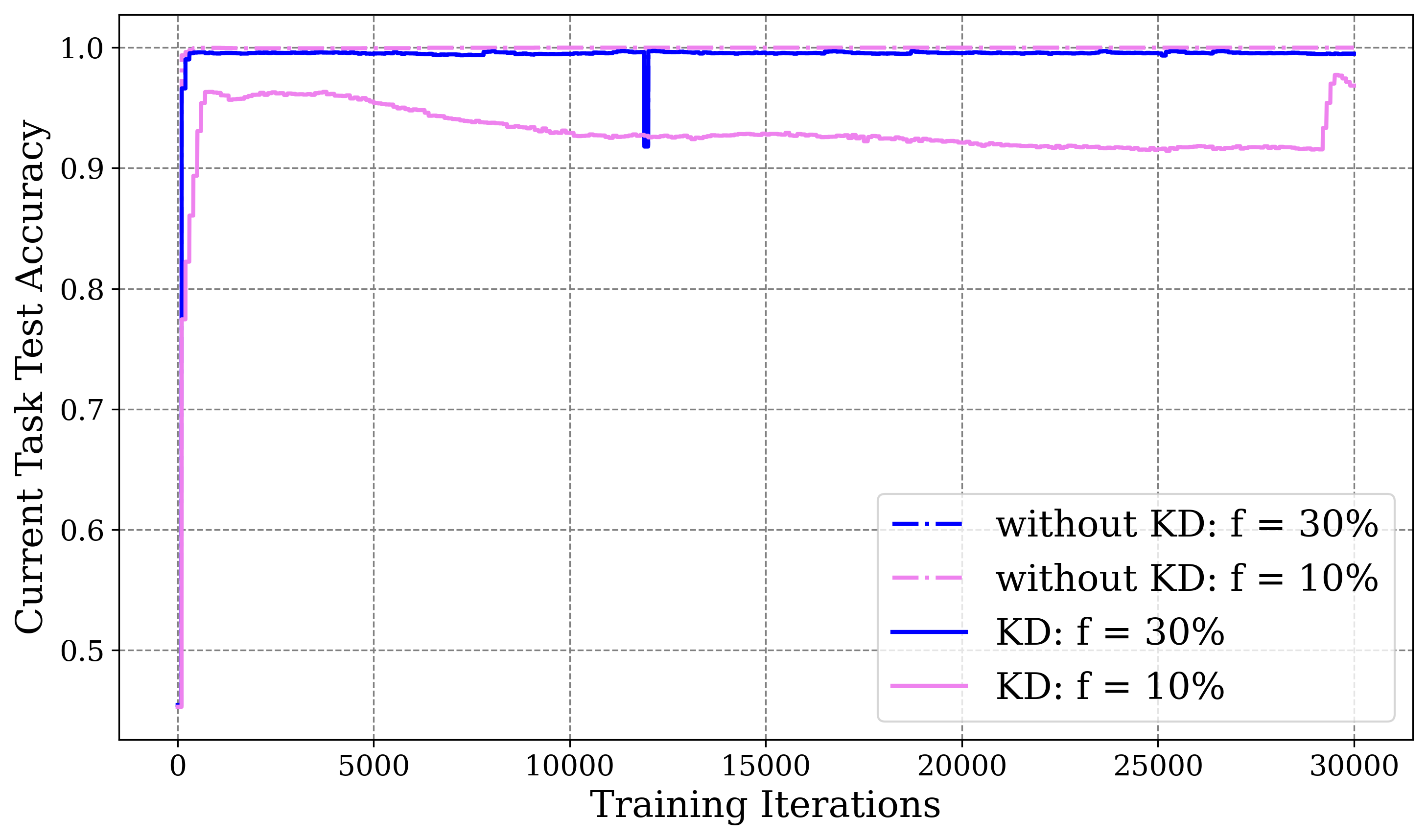}
           \caption{Current Task Accuracy for different fractions of data, with and without KD.}
           \label{fig:cta}
       \end{subfigure}
       \caption{\small Continual pretraining where a grokked model on $p_1$ is further trained on $p_2$. Without KD, performance on $p_1$ drops rapidly while generalization on $p_2$ is quick. With KD (\autoref{fig:cta}), accuracy on the current task is preserved and forgetting is mitigated. Training from a grokked model enables fast generalization without delayed grokking, though for data below the critical size we observe a sudden phase transition from $\sim92\%$ to $100\%$ accuracy at around $28K$ steps.}
       \label{fig:cl_grokking}
\end{figure}
   
% The results, as shown in \autoref{fig:cl_grokking}, demonstrate that in the absence of knowledge distillation (KD), the model experienced almost instantaneous and severe forgetting of previously acquired knowledge on $p_1$. Despite this rapid forgetting, it exhibited swift generalization to the new distribution, $p_2$. Differently, when KD was applied during continual pretraining, the model retained nearly perfect test accuracy on $p_1$ while simultaneously achieving rapid generalization on $p_2$. Notably, the incorporation of KD effectively prevented delayed generalization (grokking) in both scenarios. These findings underscore the critical role of KD in preserving previously learned information during continual pretraining. By maintaining a balance between knowledge retention and the acquisition of new skills, KD serves as a robust mechanism to enhance model stability and performance in dynamic learning environments.

As shown in \autoref{fig:cl_grokking}, training without KD led to immediate and severe forgetting of $p_1$, though the model quickly generalized to $p_2$. With KD, however, the model retained near-perfect accuracy on $p_1$ while also achieving rapid generalization on $p_2$. KD thus prevented delayed generalization and preserved prior knowledge, underscoring its role in balancing retention and adaptation during continual pretraining.

% These results underscore the ability of knowledge distillation (KD) to facilitate generalization even in scenarios where data from multiple distributions is severely limited. This is particularly relevant in practical applications where acquiring sufficient data is constrained by factors such as security protocols, privacy regulations, and other restrictions. In such contexts, leveraging KD from pre-trained grokked models presents a sophisticated and effective solution to mitigate the challenges posed by data scarcity.

These findings highlight the utility of KD in enabling generalization when data from multiple distributions is scarce, a scenario common in practice due to privacy, security, or resource constraints. Leveraging KD from pre-trained grokked models offers an effective solution in such settings. Finally, across all experiments we observed increasing weight norms despite successful grokking, challenging theories that link grokking to weight norm reduction. Instead, our results suggest that mechanisms like representation transfer via KD play a more central role, opening new directions for understanding the fundamental drivers of grokking.

% Moreover, across all our experiments, the consistent increase in weight norm—despite successful grokking—challenges existing theories that attribute grokking primarily to weight norm reduction. Our findings provide compelling evidence that alternative mechanisms, such as the transfer of learned representations via KD, play a more pivotal role in enabling generalization under limited data conditions. This insight opens new avenues for research into the fundamental drivers of grokking, moving beyond conventional optimization paradigms.

\section{Conclusions and Future Work}
   \label{sec:conclusion}
   % This study advances our understanding of the grokking phenomenon by exploring its behavior below critical data regime. Unlike prior research that primarily focused on a single training distribution and the influence of weight norm and weight decay, our work broadens the scope by systematically investigating how grokking can be induced with KD without relying only on weight decay and decreasing weight norms. Our findings challenge the prevailing notion that weight decay and decreasing weight norms are the sole drivers of grokking. Additionally, we established that KD not only accelerates the grokking process but also enables generalization below the previously identified critical data threshold even in varying distributions which is significant for scenarios characterized by data scarcity, where traditional training methods falter. 

   This study advances the understanding of grokking by examining its behavior below the critical data regime. Unlike prior work centered on single distributions and weight-norm dynamics, we show that Knowledge Distillation (KD) can induce and accelerate grokking without relying on weight decay or decreasing norms. Our results demonstrate that KD enables generalization even below the critical threshold and across varying distributions, offering a practical solution in data-scarce settings where traditional training fails. Future work may extend these insights to more complex tasks, deepen our understanding of grokking's mechanisms, and explore broader uses of pre-grokked models that generalize reliably and adapt efficiently to dynamic real-world data.

\bibliography{ref}

@article{power2022grokking,
  title={Grokking: Generalization beyond overfitting on small algorithmic datasets},
  author={Power, Alethea and Burda, Yuri and Edwards, Harri and Babuschkin, Igor and Misra, Vedant},
  journal={arXiv preprint arXiv:2201.02177},
  year={2022}
}

@article{liu2022towards,
  title={Towards understanding grokking: An effective theory of representation learning},
  author={Liu, Ziming and Kitouni, Ouail and Nolte, Niklas S and Michaud, Eric and Tegmark, Max and Williams, Mike},
  journal={Advances in Neural Information Processing Systems},
  volume={35},
  pages={34651--34663},
  year={2022}
}

@inproceedings{liu2022omnigrok,
  title={Omnigrok: Grokking beyond algorithmic data},
  author={Liu, Ziming and Michaud, Eric J and Tegmark, Max},
  booktitle={The Eleventh International Conference on Learning Representations},
  year={2022}
}

@article{humayun2024deep,
  title={Deep networks always grok and here is why},
  author={Humayun, Ahmed Imtiaz and Balestriero, Randall and Baraniuk, Richard},
  journal={arXiv preprint arXiv:2402.15555},
  year={2024}
}

@article{singh2024controlling,
  title={Controlling Forgetting with Test-Time Data in Continual Learning},
  author={Singh, Vaibhav and Aljundi, Rahaf and Belilovsky, Eugene},
  journal={arXiv preprint arXiv:2406.13653},
  year={2024}
}

@inproceedings{singh2024wake,
  title={Wake-Sleep Energy Based Models for Continual Learning},
  author={Singh, Vaibhav and Choromanska, Anna and Li, Shuang and Du, Yilun},
  booktitle={Proceedings of the IEEE/CVF Conference on Computer Vision and Pattern Recognition},
  pages={4118--4127},
  year={2024}
}

@article{nanda2023progress,
  title={Progress measures for grokking via mechanistic interpretability},
  author={Nanda, Neel and Chan, Lawrence and Lieberum, Tom and Smith, Jess and Steinhardt, Jacob},
  journal={arXiv preprint arXiv:2301.05217},
  year={2023}
}

@article{barak2022hidden,
  title={Hidden progress in deep learning: Sgd learns parities near the computational limit},
  author={Barak, Boaz and Edelman, Benjamin and Goel, Surbhi and Kakade, Sham and Malach, Eran and Zhang, Cyril},
  journal={Advances in Neural Information Processing Systems},
  volume={35},
  pages={21750--21764},
  year={2022}
}

@article{thilak2022slingshot,
  title={The slingshot mechanism: An empirical study of adaptive optimizers and the grokking phenomenon},
  author={Thilak, Vimal and Littwin, Etai and Zhai, Shuangfei and Saremi, Omid and Paiss, Roni and Susskind, Joshua},
  journal={arXiv preprint arXiv:2206.04817},
  year={2022}
}

@article{van2019three,
  title={Three scenarios for continual learning},
  author={Van de Ven, Gido M and Tolias, Andreas S},
  journal={arXiv preprint arXiv:1904.07734},
  year={2019}
}

@article{loshchilov2017decoupled,
  title={Decoupled weight decay regularization},
  author={Loshchilov, I},
  journal={arXiv preprint arXiv:1711.05101},
  year={2017}
}

@article{merrill2023tale,
  title={A tale of two circuits: Grokking as competition of sparse and dense subnetworks},
  author={Merrill, William and Tsilivis, Nikolaos and Shukla, Aman},
  journal={arXiv preprint arXiv:2303.11873},
  year={2023}
}

@InProceedings{pmlr-v139-menon21a,
  title = 	 {A statistical perspective on distillation},
  author =       {Menon, Aditya K and Rawat, Ankit Singh and Reddi, Sashank and Kim, Seungyeon and Kumar, Sanjiv},
  booktitle = 	 {Proceedings of the 38th International Conference on Machine Learning},
  pages = 	 {7632--7642},
  year = 	 {2021},
  volume = 	 {139},
  series = 	 {Proceedings of Machine Learning Research},
  month = 	 {18--24 Jul},
  publisher =    {PMLR},
 
}

@InProceedings{pmlr-v97-phuong19a,
  title = 	 {Towards Understanding Knowledge Distillation},
  author =       {Phuong, Mary and Lampert, Christoph},
  booktitle = 	 {Proceedings of the 36th International Conference on Machine Learning},
  pages = 	 {5142--5151},
  year = 	 {2019},
  volume = 	 {97},
  series = 	 {Proceedings of Machine Learning Research},
  month = 	 {09--15 Jun},
  publisher =    {PMLR},
}

@article{tang2020understanding,
  title={Understanding and improving knowledge distillation},
  author={Tang, Jiaxi and Shivanna, Rakesh and Zhao, Zhe and Lin, Dong and Singh, Anima and Chi, Ed H and Jain, Sagar},
  journal={arXiv preprint arXiv:2002.03532},
  year={2020}
}

@InProceedings{Cho_2019_ICCV,
author = {Cho, Jang Hyun and Hariharan, Bharath},
title = {On the Efficacy of Knowledge Distillation},
booktitle = {Proceedings of the IEEE/CVF International Conference on Computer Vision (ICCV)},
month = {October},
year = {2019}
}

@article{notsawo2023predicting,
  title={Predicting grokking long before it happens: A look into the loss landscape of models which grok},
  author={Notsawo Jr, Pascal and Zhou, Hattie and Pezeshki, Mohammad and Rish, Irina and Dumas, Guillaume and others},
  journal={arXiv preprint arXiv:2306.13253},
  year={2023}
}

@Article{cifar10,
author = {Krizhevsky, Alex},
year = {2012},
month = {05},
pages = {},
title = {Learning Multiple Layers of Features from Tiny Images},
journal = {University of Toronto}
}

@article{varma2023explaining,
  title={Explaining grokking through circuit efficiency},
  author={Varma, Vikrant and Shah, Rohin and Kenton, Zachary and Kram{\'a}r, J{\'a}nos and Kumar, Ramana},
  journal={arXiv preprint arXiv:2309.02390},
  year={2023}
}

@article{doshi2023grok,
  title={To grok or not to grok: Disentangling generalization and memorization on corrupted algorithmic datasets},
  author={Doshi, Darshil and Das, Aritra and He, Tianyu and Gromov, Andrey},
  journal={arXiv preprint arXiv:2310.13061},
  year={2023}
}

@article{fang2020rethinking,
  title={Rethinking importance weighting for deep learning under distribution shift},
  author={Fang, Tongtong and Lu, Nan and Niu, Gang and Sugiyama, Masashi},
  journal={Advances in neural information processing systems},
  volume={33},
  pages={11996--12007},
  year={2020}
}

@article{liang2024comprehensive,
  title={A comprehensive survey on test-time adaptation under distribution shifts},
  author={Liang, Jian and He, Ran and Tan, Tieniu},
  journal={International Journal of Computer Vision},
  pages={1--34},
  year={2024},
  publisher={Springer}
}

@inproceedings{
rubin2024grokking,
title={Grokking as a First Order Phase Transition in Two Layer Networks},
author={Noa Rubin and Inbar Seroussi and Zohar Ringel},
booktitle={The Twelfth International Conference on Learning Representations},
year={2024},
url={https://openreview.net/forum?id=3ROGsTX3IR}
}

@inproceedings{
levi2024grokking,
title={Grokking in Linear Estimators -- A Solvable Model that Groks without Understanding},
author={Noam Itzhak Levi and Alon Beck and Yohai Bar-Sinai},
booktitle={The Twelfth International Conference on Learning Representations},
year={2024},
url={https://openreview.net/forum?id=GH2LYb9XV0}
}

@inproceedings{
kumar2024grokking,
title={Grokking as the transition from lazy to rich training dynamics},
author={Tanishq Kumar and Blake Bordelon and Samuel J. Gershman and Cengiz Pehlevan},
booktitle={The Twelfth International Conference on Learning Representations},
year={2024},
url={https://openreview.net/forum?id=vt5mnLVIVo}
}

@inproceedings{
lyu2024dichotomy,
title={Dichotomy of Early and Late Phase Implicit Biases Can Provably Induce Grokking},
author={Kaifeng Lyu and Jikai Jin and Zhiyuan Li and Simon Shaolei Du and Jason D. Lee and Wei Hu},
booktitle={The Twelfth International Conference on Learning Representations},
year={2024},
url={https://openreview.net/forum?id=XsHqr9dEGH}
}

@inproceedings{Vaswani2017AttentionIA,
  title={Attention is All you Need},
  author={Ashish Vaswani and Noam M. Shazeer and Niki Parmar and Jakob Uszkoreit and Llion Jones and Aidan N. Gomez and Lukasz Kaiser and Illia Polosukhin},
  booktitle={Neural Information Processing Systems},
  year={2017},
  url={https://api.semanticscholar.org/CorpusID:13756489}
}

@InProceedings{Yuan_2020_CVPR,
author = {Yuan, Li and Tay, Francis EH and Li, Guilin and Wang, Tao and Feng, Jiashi},
title = {Revisiting Knowledge Distillation via Label Smoothing Regularization},
booktitle = {Proceedings of the IEEE/CVF Conference on Computer Vision and Pattern Recognition (CVPR)},
month = {June},
year = {2020}
}

@inproceedings{ke2022dgs,  
 title={Continual Learning of Language Models}, author={Ke, Zixuan and Shao, Yijia and Lin, Haowei and Konishi, Tatsuya and Kim, Gyuhak and Liu, Bing}, booktitle={International Conference on Learning Representations (ICLR)}, year={2023}}

@inproceedings{Arpit2017ACL,
  title={A Closer Look at Memorization in Deep Networks},
  author={Devansh Arpit and Stanislaw Jastrzebski and Nicolas Ballas and David Krueger and Emmanuel Bengio and Maxinder S. Kanwal and Tegan Maharaj and Asja Fischer and Aaron C. Courville and Yoshua Bengio and Simon Lacoste-Julien},
  booktitle={International Conference on Machine Learning},
  year={2017},
}

@article{Ishida2020DoWN,
  title={Do We Need Zero Training Loss After Achieving Zero Training Error?},
  author={Takashi Ishida and Ikko Yamane and Tomoya Sakai and Gang Niu and Masashi Sugiyama},
  journal={ArXiv},
  year={2020},
  volume={abs/2002.08709},
  url={https://api.semanticscholar.org/CorpusID:211205200}
}

@misc{boixadsera2024theorymodeldistillation,
      title={Towards a theory of model distillation}, 
      author={Enric Boix-Adsera},
      year={2024},
      eprint={2403.09053},
      archivePrefix={arXiv},
      primaryClass={cs.LG},
      url={https://arxiv.org/abs/2403.09053}, 
}

@inproceedings{ben2011learning,
  title={Learning a classifier when the labeling is known},
  author={Ben-David, Shalev and Ben-David, Shai},
  booktitle={International Conference on Algorithmic Learning Theory},
  pages={440--451},
  year={2011},
  organization={Springer}
}

@inproceedings{
prieto2025grokking,
title={Grokking at the Edge of Numerical Stability},
author={Lucas Prieto and Melih Barsbey and Pedro A. M. Mediano and Tolga Birdal},
booktitle={The Thirteenth International Conference on Learning Representations},
year={2025},
url={https://openreview.net/forum?id=TvfkSyHZRA}
}

@article{critical_data,
  author={Xuekai Zhu and Yao Fu and Bowen Zhou and Zhouhan Lin},
  title={Critical Data Size of Language Models from a Grokking Perspective},
  year={2024},
  cdate={1704067200000},
  journal={CoRR},
  volume={abs/2401.10463},
  url={https://doi.org/10.48550/arXiv.2401.10463},
}

@inproceedings{xu2025let,
title={Let Me Grok for You: Accelerating Grokking via Embedding Transfer from a Weaker Model},
author={Zhiwei Xu and Zhiyu Ni and Yixin Wang and Wei Hu},
booktitle={The Thirteenth International Conference on Learning Representations},
year={2025},
url={https://openreview.net/forum?id=4rEI2JdHH6}
}

@article{lee2024grokfast,
  title={Grokfast: Accelerated Grokking by Amplifying Slow Gradients},
  author={Lee, Jaerin and Kang, Bong Gyun and Kim, Kihoon and Lee, Kyoung Mu},
  journal={arXiv preprint arXiv:2405.20233},
  year={2024}
}

@article{minegishi2023bridging,
  title={Bridging Lottery ticket and Grokking: Is Weight Norm Sufficient to Explain Delayed Generalization?},
  author={Minegishi, Gouki and Iwasawa, Yusuke and Matsuo, Yutaka},
  journal={arXiv preprint arXiv:2310.19470},
  year={2023}
}
\bibliographystyle{plain}
\newpage
\appendix
\section{Appendix}
\subsection{Evolution of L2 weight norm}
\label{appendix:l2norm}
\begin{figure}[ht]
    \centering
   \includegraphics[width=0.75\textwidth]{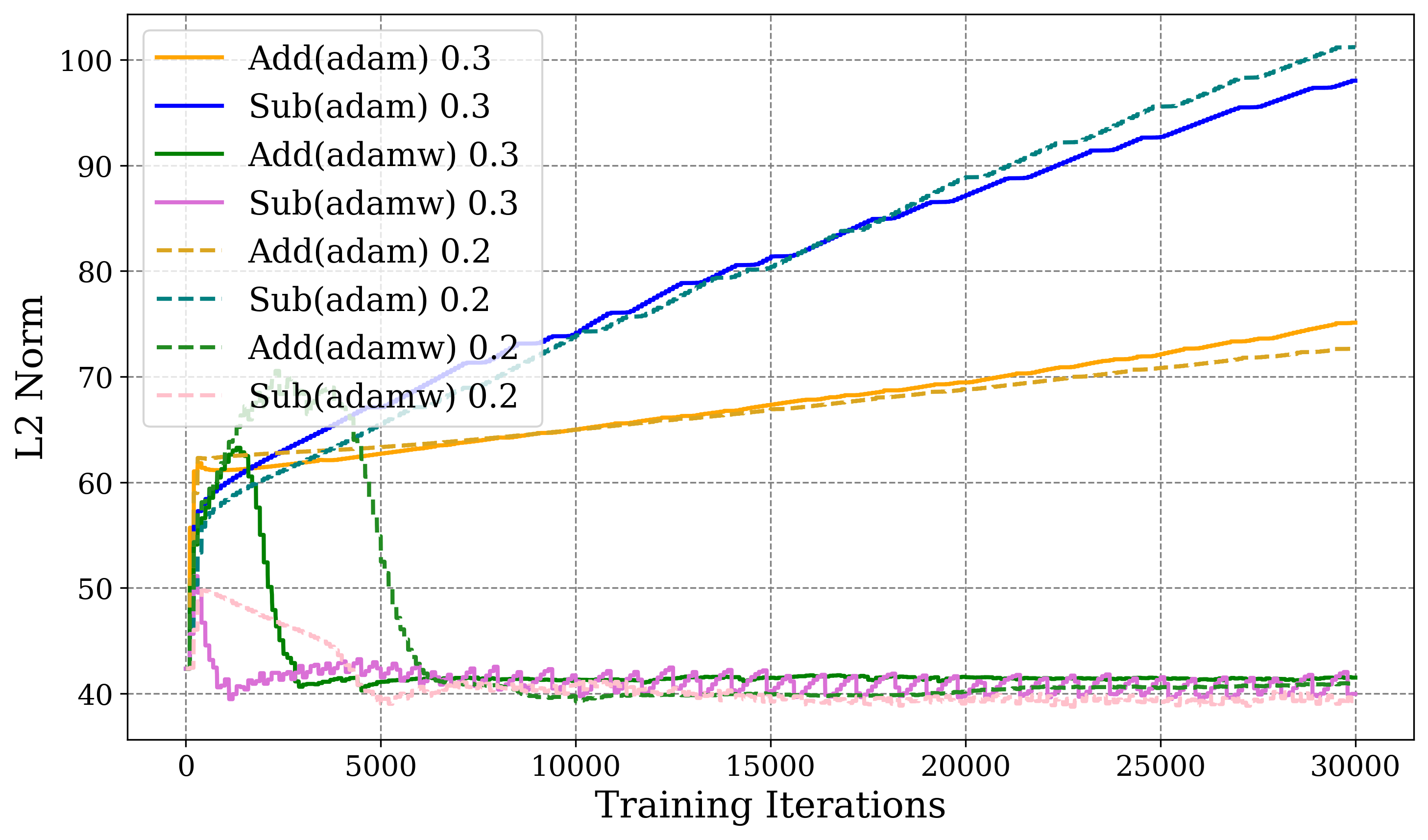}
    \caption{Evolution of the L2 weight norm for $Student$ model \fs\ trained with Adam(without weight decay) and AdamW(with weight decay) on different fractions of $p_2$ distribution. \fs\ is trained via KD from a grokked model \ft\ . Notably, training without weight decay the $L_2$-weight norm increases continuously, while giving generalised solutions. This rules out the necessity of decreased weight norm condition for exhibiting grokking given by \citep{liu2022omnigrok, varma2023explaining, nanda2023progress}}
     \label{fig:norm}
  \end{figure}

L2 weight norm trained with adam continuously increases for both addition and subtraction tasks as observed in \autoref{fig:norm}, yet grokking still occurs. These findings challenge the theories proposed by~\citep{nanda2023progress} who suggest that the abrupt transition to perfect test accuracy during grokking occurs in the cleanup phase (where weight decay removes memorization components), following the establishment of the generalizing mechanism. Our empirical evidence contradicts these claims by demonstrating grokking even without weight decay and with increasing weight norms.
 
Similarly \cite{liu2022omnigrok} induce grokking by increasing the initial weight norm and conclude that generalizing solutions lie on smaller norm spheres in parameter space. While we acknowledge that an initially higher weight norm can facilitate grokking, our results indicate that generalizing solutions do not necessarily lie on smaller norm spheres. Our modular arithmetic tasks serve as counterexamples, where the final generalizing solutions exhibit larger parameter weight norms than their initial states, and grokking occurs without the application of weight decay.

Furthermore~\cite{varma2023explaining} claim that the transition from memorizing to generalizing circuits occurs because the generalizing circuit is more “efficient” than the memorizing circuit, in the sense that it can produce equivalent loss with a lower parameter norm. In contrast, our studies show that modular arithmetic tasks can achieve generalizing solutions with higher parameter norms without any weight decay, disproving the necessity of norm reduction for grokking. 

We empirically demonstrate that parameters' (L2) weight norm  trained with adam continuously increases for both addition and subtraction tasks as observed in \autoref{fig:norm}, yet grokking still occurs. This challenges the notion that decreasing weight norm is fundamental to grokking. Therefore, we assert that neither parameter weight decay nor a decreasing weight norm during optimization is inherently fundamental to observing grokking, contrary to its purported necessity in previous studies \citep{liu2022omnigrok, nanda2023progress, varma2023explaining} on modular arithmetic tasks.

% \subsection{Distilling from multiple grokked models}
%   \begin{figure}[ht]
%      \centering
%     \includegraphics[width=0.55\textwidth]{images/stable_loss/kd_multiple.png}
%      \caption{\small Training of a larger model \fm\ via distilling from grokked models $f_{p_1}$ and $f_{p_2}$.These small models are grokked on $35\%$ of training data each. Training of larger model \fm\ is trained with different fractions($0.35, 0.3, 0.25$) of $p_1$ and $p_2$, with only distillation from grokked models $f_{p_1}$ and $f_{p_2}$. }
%       \label{fig:kd_multiple_orig}
%    \end{figure}

\end{document}